\documentclass[preprint,12pt]{elsarticle}

\usepackage{graphicx}
\usepackage{natbib}
\usepackage{subcaption}
\usepackage{multirow}
\usepackage{booktabs}
\usepackage{float}
\usepackage{tablefootnote}
\usepackage{xcolor}
\usepackage{hyperref}
\usepackage{caption}
\usepackage{changepage}
\usepackage{makecell}
\usepackage{tabularray}
\usepackage[justification=centering]{caption}
\usepackage{placeins}
\usepackage{amssymb}
\usepackage{amsmath}

\journal{Natural Language Processing}

\begin{document}

\begin{frontmatter}



\title{A Multi-way Parallel Named Entity Annotated Corpus for English, Tamil and Sinhala}


\author[label1]{Surangika Ranathunga\corref{cor1}}
\ead{s.ranathunga@massey.ac.nz}
\cortext[cor1]{Corresponding author}
\author[label2]{Asanka Ranasinghe}
\ead{asankar.17@cse.mrt.ac.lk}
\author[label2]{Janaka Shamal}
\ead{janakashamal.17@cse.mrt.ac.lk}
\author[label2]{Ayodya Dandeniya }
\ead{ayodya.18@cse.mrt.ac.lk}
\author[label2]{Rashmi Galappaththi}
\ead{rashmi.18@cse.mrt.ac.lk}
\author[label2]{Malithi Samaraweera}
\ead{malithi.18@cse.mrt.ac.lk}
\affiliation[label1]{organization={School of Mathematical and Computational Sciences},
            addressline={Massey University}, 
            city={Auckland},
            postcode={102904}, 
            country={New Zealand}}    
\affiliation[label2]{organization={Department of Computer Science and Engineering},
            addressline={University of Moratuwa}, 
            city={Katubedda},
            postcode={10400}, 
            country={Sri Lanka}}

\begin{abstract}
This paper presents a multi-way parallel English-Tamil-Sinhala corpus annotated with Named Entities (NEs), where Sinhala and Tamil are low-resource languages. Using pre-trained multilingual Language Models (mLMs), we establish new benchmark Named Entity Recognition (NER) results on this dataset for Sinhala and Tamil. We also carry out a detailed investigation on the NER capabilities of different types of mLMs. Finally, we demonstrate the utility of our NER system on a low-resource Neural Machine Translation (NMT) task. Our dataset is publicly released: \url{https://github.com/suralk/multiNER}.
\end{abstract}




\end{frontmatter}



\section{Introduction}
Named  Entity Recognition (NER) is the process of identifying Named Entities (NEs) in natural language text. An NE can be a word or a phrase, and the detected entities are categorized into predetermined categories such as person, location and organization. As an example, consider the sentence: \textit{John works in Facebook at Los Angeles}. This sentence contains three NEs: John (person), Facebook (organization) and Los Angeles (location). NER either acts as an intermediate step for, or helps to improve many high-level Natural Language Processing (NLP) tasks such as question answering \citep{lamurias2019lasigebiotm}, Neural Machine Translation (NMT)~\citep{hu2021deep}, Information Retrieval~\citep{guo2009named} and automatic text summarization \citep{khademi2020persian}. There have been recent developments with respect to NE tag sets \citep{bib1}, tagging schemes \citep{bib3}, as well as NER algorithms \citep{bib1}. Most of the NER algorithms that produced promising results are supervised, meaning that they are trained with NE annotated datasets. However, many low-resource languages have little to no annotated data~\citep{joshi2020state}.

Sinhala and Tamil are examples of low-resource languages~\citep{joshi2020state, ranathunga2022some} with limited NE annotated data~\citep{de2019survey}. Manamini et al.'s~\cite{bib19} is the only publicly available Sinhala NE dataset that has been manually annotated. FIRE corpus\footnote{\url{http://fire.irsi.res.in/fire/2023/home}} is the only publicly available manually annotated NE dataset for Tamil. WikiANN \citep{pan2017cross}  and LORELEI\citep{tracey2020basic} are multilingual (but not multi-way pararell) NE annotated datasets that contain both Sinhala and Tamil. However, WikiANN has been automtically annotated using entity linking and LORELEI is hidden behind a paywall. 

In this paper, we present a multi-way parallel English-Sinhala-Tamil NE annotated corpus that consists of 3835 sentences per language. This corpus is annotated using the  CONLL03 tag set \citep{bib10}, which has four tags: persons (PER), locations (LOC), organizations (ORG) and miscellaneous (MISC). The corpus was annotated using Beginning-Inside-Outside (BIO) format. We provide a comprehensive analysis of this dataset and the data creation process. This dataset is publicly released\footnote{\url{https://github.com/suralk/multiNER}}.

The benefit of having such multi-way parallel datasets is that they can serve as good test beds to evaluate language-specific NER capabilities of pre-trained multilingual Language Models (mLMs) such as mBERT~\citep{devlin2018bert} and XLM-R~\citep{bib17}, which form the basis of modern-day NLP systems. In other words, if the mLM is fine-tuned with data from individual languages, this results in language-specific NER models, which can be probed to identify how their performance varies across languages. On the other hand, fine-tuning the mLM with all the language data of the multi-way parallel corpus results in a single NER model that caters for all the languages included in the corpus. How the performance of these models varies depending on the linguistic properties of individual languages is useful in building optimal language-specific NER systems.

Despite the NER models built on mLMs outperforming the more traditional Deep Learning (DL) models such as Recurrent Neural Networks (specifically BiLSTM-CRF (Bi-Directional Long Short Term Memory with a CRF layer)~\citep{bib1, yadav2018deep}), these newer techniques have not been employed for Sinhala or Tamil NER.

In order to evaluate the Sinhala and Tamil NER capabilities, we experimented with different types of pre-trained LMs (pLMs): language-specific pLMs, language family-specific pLMs and mLMs. Our results show that, when the language is already included in the pLM, NER systems built on all these type of pLMs significantly outperform those that use the BiLSTM-CRF model. We also show that a multilingual NER model built by fine-tuning XLM-R with our multi-way parallel corpus outperforms (or is on-par with) the NER models trained for individual languages. Finally, we demonstrate the utility of the built NER models by using their output in building an NMT system. The NMT system, which was trained using the NEs identified using our NER system as an additional input significantly outperformed the baseline NMT model.

\section{Related Work}
\label{section:rel}
In this section, we provide a brief overview of NE tag sets, annotation schemes, NE annotated datasets and NER techniques based on pLMs. We also discuss NER research available for Sinhala and Tamil.

\subsection{NE Tag Sets}
 One prominent tag set that has been widely used in NER is the CONLL03 tag set \citep{bib10}. This tag set has only 4 tags - Person, Location, Organization and Miscellaneous.  The
Co-reference and Entity Type tag set \citep{bib12} has 12 NE types (7 numerical types, 10 nominal entity types and temporal types). WNUT2017~\citep{derczynski2017results} corpus has 6 entity types. ACE 2005/2008 \citep{bib11} has identified 7 NE types, as
people, organizations, locations, facilities, Geo political entities, weapons, vehicles and events. ACE was defined for the tourism domain, thus can be considered as a domain-specific tag set.
While having more NE types helps in extracting more information, training Machine Learning (ML) models for the task becomes very data intensive, which in turn makes manual data annotation costly. As shown by Azeez and Ranathunga~\cite{bib18}, if a sufficient annotated corpus is not created, having a fine-grained tag set is not useful, as many NE tags will not have sufficient data points to train the ML model.

\subsection{Annotation Schemes}
Flat NE annotation schemes have been most commonly used in previous research. Alshammari et al.~\cite{alshammari2021impact} present a comprehensive list of such annotation schemes. Some popular NE annotation schemes are IO (tags each token either as an inside tag (I) or an outside tag (O), where NEs are marked as I), IOB (which is also known as BIO - tags each token either as beginning (B) of a known NE, inside (I) it, or outside (O) of any known NE) and IOE (similar to IOB, but tags the end of an NE (E) instead of its beginning).

The main drawback of these flat annotation schemes is that they are unable to capture nested NEs. As an alternative, markup NE annotation schemes have been proposed~\citep{mitchell2005ace, ringland2019nne, marcus2011ontonotes}.
While more comprehensive annotation schemes allow the extraction of more fine-grained information, this requires more input from humans who annotate the data and ML models need more data to learn.  Therefore, still the BIO tagging scheme, which was used in the CONLL 2003 shared task~\citep{bib10} is being commonly used.

\subsection{NER techniques}
\label{section:rel_techniques}
Recently, pLMs such as BERT~\citep{devlin2018bert}, RoBERTa ~\citep{liu2019roberta} and their multilingual variants such as mBERT and XLM-R are widely used for the NER task. In fact, NER was one of the evaluation tasks used for the BERT and XLM-R papers. Subsequent research has introduced various improvements on the vanilla fine-tuning of pLMs for NER~\citep{souza2019portuguese, fetahu2022dynamic, huang2022copner, fu2022cross}. This line of research has outperformed the traditional DL techniques such as RNNs by a significant margin. Therefore here we discuss only NER techniques based on pLMs. For a comprehensive survey on the previous DL techniques for NER, we refer the reader to Yadav et al. \cite{yadav2019survey}.

Some research combined NE corpora from multiple languages (not multi-way parallel) to build a single multilingual NER model on top of a pre-trained multilingual language model~\citep{kulkarni2023towards,shaffer2021language}, which has shown to perform better than language-specific models. This is due to the cross-lingual transfer of knowledge across languages in the model. Thus such multilingual models are a very promising solution for low-resource languages. If there is no NE annotated data for a target language, one solution is to synthetically generate NE data using an NER model of a high-resource language~\citep{li2021cross, yang2022crop}. Another popular solution is Transfer Learning, where an NER model fine-tuned on an pLM is used to infer NE information on a target language~\citep{bib9}.

All Sinhala NER research except Rijhwani et al.~\citep{rijhwani2020soft} who implemented an LSTM based NER model for Sinhala is based on traditional Machine Learning (ML) models such as SVM and CRF~\citep{dahanayaka2014named, bib19, bib24, senevirathne2015conditional, wijesinghe2022sinhala}. Similarly for Tamil, except Anbukkarasi et al.~\cite{anbukkarasi2022named} and Hariharan et al.~\cite{hariharan2019named}, all the other research is based on traditional ML techniques~\citep{ murugathas2022domain,srinivasagan2014automated, srinivasan2019automated, abinaya2015randomized, vijayakrishna2008domain, antony2014named, abinaya2014amrita_cen, gayen2014hmm, murugathas2022domain, theivendiram2018named, mahalakshmi2016domain, malarkodi2020deeper, malarkodi2012tamil, ram2010linguistic}.

\subsection{NE Annotated Corpora for Sinhala and Tamil}
Manamini et al.~\cite{bib19} produced an NE annotated dataset\footnote{\url{http://bit.ly/2XrwCoK}} for Sinhala. However, it has not used a standard tagging schema such as BIO. Other Sinhala NER research has not publicly released their datasets. The FIRE corpus is the most commonly used dataset for Tamil NER. Other research~\citep{antony2014named} that created NER datasets has not published their data. 

WikiAnn corpus \citep{pan2017cross} is one of the efforts to build an NE annotated dataset for low resource languages. It was created by transferring annotations from English to other 282 languages
through cross-lingual links in knowledge bases. Thus, the WikiAnn corpus is said to have “silver-standard” labels. Although it has both Sinhala and Tamil data, the amount is rather small, specially for Sinhala. Moreover, WikiAnn corpus  has been annotated with only three major entities: Person, Organization and Location. LORELEI~\citep{tracey2020basic} is another multilingual NE annotated corpus that includes both Sinhala and Tamil. Unlike WikiAnn, LORELEI is a manually annotated corpus. However, it is hidden behind a paywall.

Although Sinhala or Tamil are not included, two other notable multilingual corpora are 
MultiCoNER~\citep{malmasi2022multiconer} and MasakhaNER~\citep{bib9}. MultiCoNER was built using a technique similar to WikiANN. However, data for low-resource languages has been created by Machine Translation. MasakhaNER corpus contains manually annotated data for low-resource African languages. None of these are multi-way parallel.\\

\section{Multi-way Parallel English-Tamil-Sinhala Dataset}
\label{section:data}
In this section, first we give a brief introduction to Sinhala and Tamil. Then we discuss the data source used to create the NE annotated dataset, the pre-processing and annotation steps we have performed, data format we used and statistics of the multi-way parallel dataset.

\subsection{Sinhala}
Sinhala is an Indo-Aryan language spoken by about 16 million people, primarily in the island nation Sri Lanka. Sinhala has its own alphabet and script. In Ranathunga and de Silva's~\cite{ranathunga2022some} language categorisation, Sinhala is categorized into class 2. According to Joshi et al.'s~\cite{joshi2020state} language category definition,  this means Sinhala has only a small amount of annotated datasets.

\subsection{Tamil}
Tamil is a Dravidian language spoken by about 78 million people, primarily in the Indian state of Tamil Nadu. Tamil is also spoken by a significant population mainly in Sri Lanka, Singapore and Malaysia. Tamil is written in the Tamil script, which is an Abugida (syllabic) script. In Ranathunga and de Silva's~\cite{ranathunga2022some} language categorisation, Tamil is categorised into class 3, meaning that it has small amounts of labelled datasets, but with a better web presence than Sinhala.

\subsection{Raw Data}

Our data comes from the multi-way parallel English-Tamil-Sinhala dataset developed by Fernando et al.~\cite{fernando2020data}. This corpus contains official government documents, namely annual reports, letters and circulars. Despite being specific to the government domain, this dataset has a wide coverage mainly because of the inclusion of annual reports coming from  different
government institutions corresponding to Art, Media, Finance, Education, Technology, Procurement, etc.\\
This corpus contains duplicate sentences, unwanted long lists (with 200+ tokens per list) such as table of contents and meaningless sentences. Thus, we manually filtered 3835 Sinhala sentences\footnote{We started with Sinhala sentences because Fernando et al.'s~\cite{fernando2020data} corpus was compiled by taking Sinhala as the source.}. During the filtering process, we removed meaningless sentences, duplicate sentences, undesirable lists and captions/ headers of figures/tables from the dataset. For each Sinhala sentence  in this filtered corpus, we extracted the corresponding Tamil and English sentences from Fernando et al.~\cite{fernando2020data}'s raw parallel corpus.

\subsection{Data Annotation}
\label{sect:Test_dataset}
\begin{figure*}[!h]
\begin{center}
\includegraphics[scale=0.8]{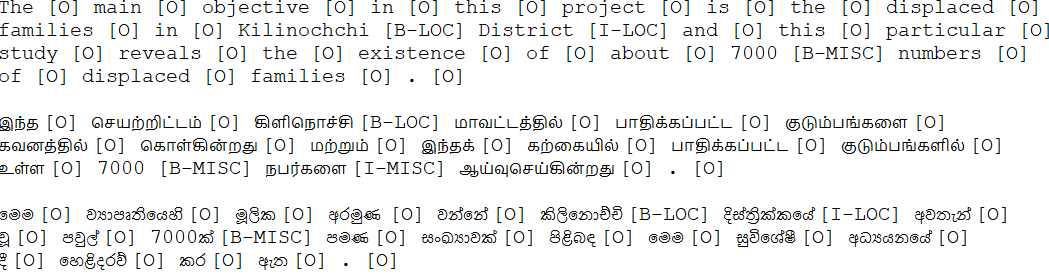} 
\caption{Sample English/Tamil/Sinhala  sentences annotated with CONLL03 tag set, following the BIO format.}
\label{fig.1}
\end{center}
\end{figure*}

Since our dataset has only about 100k tokens per language, we did not want to go for a fine-grained NE tag set. Therefore the annotation was carried out using the CONLL03 tag set. BIO annotation scheme was used for annotation. This corpus was annotated manually using the Inception annotation tool\footnote{\url{https://inception-project.github.io/downloads/}}. Two independent annotators for each language were employed to annotate the dataset. Annotators were  provided an in-house training and annotation guidelines. Later, in order to establish the inter-annotator agreement, two more annotators annotated about 500 tokens from each language. The inter-annotator agreement values were reported as 0.83, 0.89 and 0.88 for Sinhala, English and Tamil (respectively).\\
Fig.~\ref{fig.1} is an example annotated sentence in all three languages.

\subsection{Data Statistics}
Our final dataset consists of 3835 parallel sentences per language.  Sinhala, Tamil and English vocabularies of the prepared dataset contains 11560, 19308 and 10607 distinct word tokens respectively. Table \ref{tab:tb2}  shows the tag counts in each dataset. Table~\ref{tab:tb3} shows the entity percentage in each dataset. Altogether, 9.74\% of tokens in the corpus are NEs. This amount is in a similar range as the amount of NEs identified in the CoNLL03 dataset. Given that this dataset is from official government documents, it has more location and organization NEs than person NEs. The percentage of all the other NEs being just 6.55\% justifies our selection of the ConLL tagset with 4 tags - Had the Miscellaneous tag been expanded into unique NEs, each category would have a very low amount of samples annotated.
\begin{table}
    \caption{Number of entities in each  dataset}
    \centering
    \begin{tabular}{|c| c | c |c|} 
 \hline
 \bf{Tag} &  \bf{English} & \bf{Sinhala} & \bf{Tamil}\\ [0.5ex] 
 \hline
  B-PER & 194 & 194 & 226 \\
        \hline
        I-PER & 542 & 610 & 406\\
        \hline
        B-ORG & 1539 & 1628 & 1420\\
        \hline
        I-ORG & 3721 & 3666 & 2247\\
        \hline
        B-LOC & 1511 & 1728 & 1829\\
        \hline
        I-LOC & 552 & 780 & 437
        \\
        \hline
        B-MISC & 6283 & 6549 & 6012
        \\
        \hline
        I-MISC & 8587 & 7354 & 7543
        \\
        \hline
        O & 82335 & 73493 & 62947
        \\
        \hline
\end{tabular}

    \label{tab:tb2}
\end{table}

According to Table \ref{tab:tb2}, we could see that B-tag counts of Person, Organization and Location entities do not tally up across languages. Following are the reasons (and an example for each) we identified for this discrepancy. 

\begin{flushleft}

\begin{enumerate}
   \item Translation discrepancies due to dissimilar language syntax
  
        \begin{figure}[h]
\begin{center}
\includegraphics[scale=0.8]{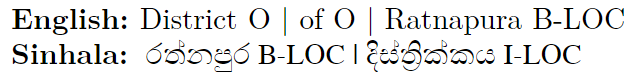} 
\end{center}
\end{figure}
        Here, \textit{Rathnapura District}, has been wriitten as  \textit{District of Rathnapura}, which has been marked as having one NE. However the Sinhala translation of it contains two NE tags.
     
   \item Differences in how abbreviations and acronyms have been handled in different languages
    \begin{figure}[h]
\begin{center}
\includegraphics[scale=0.8]{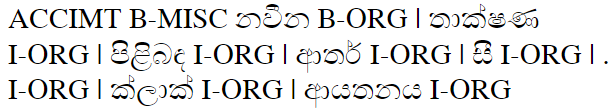} 
\end{center}
\end{figure}   
   
    The English sentence has the NE as an acronym, whereas the Sinhala version has the expanded version of it. Note that the English abbreviation ACCIMT has been annotated as B-MISC, while the corresponding Sinhala entity has been identified as an ORG.

 \item Annotation mistakes by human annotators\\
    \begin{figure}[h]
\begin{center}
\includegraphics[scale=0.8]{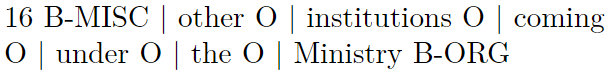} 
\end{center}
\end{figure}
    
    In this phrase, the last term has been annotated as B-ORG simply because the word (incorrectly) starts with capitalization, although it does not refer to any specific ministry.
    
\end{enumerate}

\end{flushleft}

\section{Methodology}
As mentioned in Section~\ref{section:rel_techniques}, the DL techniques used so far for Sinhala and Tamil NER are based on RNN models. Therefore we used one of these models as the baseline. To be specific, we used a Bi-LSTM CRF model, as it has been the state-of-the-art architecture before the introduction of pre-trained pLMs~\citep{yadav2019survey}.

\begin{table}[hbt!]
    \caption{ Percentage of NEs in each dataset}
     \centering
     \begin{tabular}{|c| c | c |c|} 
     \hline
 \bf{NE Type} &
        \bf{English} &
        \bf{Sinhala} &
        \bf{Tamil}\\
        \hline
        PER & 0.18 & 0.19 & 0.26
        \\
        \hline
        LOC & 1.38 & 1.73 & 2.10
        \\
        \hline
        ORG & 1.63 & 1.41 & 1.63
        \\
        \hline
        MISC & 6.55 & 5.76 & 6.92 \\
        \hline
 \end{tabular}

    \label{tab:tb3}
\end{table}

As for NER model implementation with pLMs, we experimented with three variants: a language-specific pLM, a language-family specific pLM, and two mLMs.  

\subsection{Bi-LSTM CRF method}
\begin{figure*}[!h]
\begin{center}
\includegraphics[width=\textwidth]{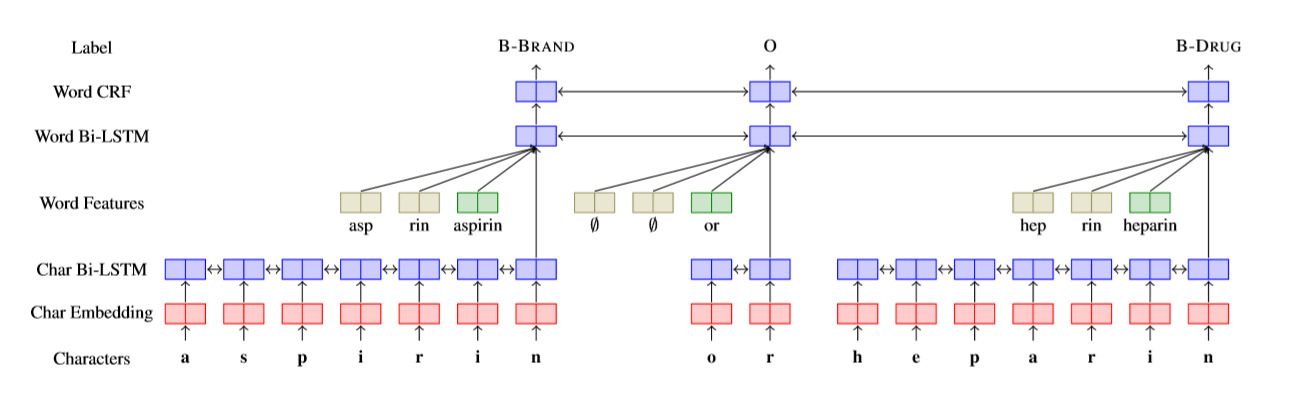} 
\caption{Architecture of the Bi-LSTM CRF network with affix features \citep{yadav2018deep} }
\label{fig.3}
\end{center}
\end{figure*}

We used the BiLSTM-CRF model used by Yadav et al.~\cite{yadav2018deep}, which has shown to be the best recurrent model for NER.  In addition to character and word embeddings, this model uses prefix-suffix embeddings as the input. The model architecture is shown in Fig.~\ref{fig.3}. In our implementation, we used FastText to obtain word embeddings. We experimented with both prefix and suffix features, however only prefix features resulted in improved results, hence only this result is reported. We used three techniques to generate character embeddings: 

\begin{enumerate}
   \item Convolutional Neural Networks (CNNs)~\citep{ma2016end}
   \item LSTMs~\citep{lample2016neural}
   \item LSTM for character embedding + Affix features\citep{yadav2018deep}  
\end{enumerate}

\subsection{ Fine tuning pre-trained pLMs}
\textbf{mLMs: } We experimented with mBERT and XLM-R. Both mBERT and XLM-R have been trained with the Masked Language Model (MLM) objective. However the creators of XLM-R claim that it is better tuned than XLM-R, and the same is confirmed by their experiment results. XLM-R has been trained with 100 languages, while mBERT has been trained with 104 languages. Tamil is included in both these models, however Sinhala is included only in XLM-R.

\textbf{Language family-specific pLMs}: We experimented with IndicBERT~\citep{kakwani2020indicnlpsuite}, which has been specifically trained for 12 Indian languages, primarily belonging to Indo-Aryan and Dravidian language families. Tamil is included in indicBERT. Despite being an Indo-Aryan language, Sinhala is not included in this model.

\textbf{Language-specific pLMs:} SinBERT~\citep{dhananjaya2022bertifying} was used as the language-specific pLM for Sinhala. SinBERT was built following RoBERTa. The dataset used to train SinBERT comprises of 15.7 million Sinhala sentences. Note that there is no Tamil-specific pLM.

Language coverage of each of these models is shown in Table~\ref{tab:LLMs}. Each of these pLMs was separately fine-tuned with our annotated data from individual languages for the NER task. A feed-forward layer on top of the pLM was used for classification. As for the multilingual model, we concatenated all Sinhala, English and Tamil datasets and sampled them with a fraction of 1 to mix sentences of the three languages. 
\begin{table}
\caption{Language Coverage in Different pLMs}
\label{tab:LLMs}
\centering
\begin{tabular}{|c| c | c |c|c|} 
 \hline
 {} & mBERT & XLM-R & IndicBERT & SinBERT\\ [0.5ex] 
 \hline
 Sinhala & {}  & \checkmark & {} & \checkmark \\ 
 \hline
 Tamil & \checkmark & \checkmark & \checkmark & {}\\ 
 \hline
 English & \checkmark & \checkmark & \checkmark & {} \\ 
 \hline
\end{tabular}
\end{table}

\section{Experiment Setup}
\label{section:experiment_setup}
For experiments, we selected XLM-R base, mBERT base, SinBERT small and IndicBERT models. Hyperparameter tuning was done using Optuna Python library\footnote{\url{https://optuna.org/}}.  For all the models, training batch size 8, evaluation batch size 16, 3 training epochs and a weight decay of 0.01 gave the optimal results. As shown in Table \ref{tab:tb3}, only the learning rate was varied.

\begin{table}[t]
    \caption{Best hyper-parameter values}
    \label{tab:tb4}

    \centering
    \begin{tabular}{|c| c | } 
    \hline
    \bf{Model} &
        \bf{Learning rate} \\
        \hline
        XLM-R base(Si, Ta, En) &
        3e-5 \\
        \hline
        mBERT base(Si, Ta, En) &
        3e-5 \\
        \hline
        SinBERT small(Si) &
        2.5e-4 \\
        \hline
        SinBERT small(Ta, En) &
        2.5e-5 \\
        \hline
        IndicBERT(Si, En) &
        9e-5 \\
        \hline
        IndicBERT(Ta) &
        5e-5 \\
        \hline
\end{tabular}

\end{table}

Evaluations were done according to the holdout method. The dataset was divided as 70-10-20 for train/validation/test sets, respectively. Each experiment was run three time with three different seeds (22, 42, 62) and then the average was calculated to get the final results. 

All experiments were carried out using the Google Colab platform under the following GPU configurations: CUDNN Version - 8700, Number CUDA Devices - 1, CUDA Device Name - Tesla T4, CUDA Device Total Memory [GB] - 15.8.

\section{Evaluation}
\newcommand{\cl}[1]{\textcolor{lightgray}{#1}}
\begin{table}
    \caption{Macro F1 score for different models. L- LSTM for char embeddings, C- CNN for char embeddings, L+A - LSTM for char embeddings with affix features}
    \label{tab:tb5}

     \centering
     \resizebox{\textwidth}{!}{
 \begin{tabular}{|c| c | c |c|c|c|c|c|c|} 
  \hline
        \bf{} &
        \bf{BiLSTM CRF (L)} &
        \bf{BiLSTM CRF (C)} & 
        \bf{BiLSTM CRF (L+A)} &
        \bf{mBERT} &
        \bf{XLM-R} &
        \bf{mXLM-R} &
        \textbf{IndicBERT} &
        \textbf{SinBERT}\\
        \hline
        Sinhala &
        64.91 & 
        64.67 &
        65.66 &
        \cl{60.53} &
        87.71 &
       \textbf{ 88.33 }&
        \cl{49.59} &
        83.77\\
        \hline
        Tamil &
        46.64 & 
        42.20 &
        47.19 &
        77.46 &
        78.81 &
        \textbf{80.23} &
        65.93 &
        \cl{41.40}\\
        \hline
        English &
        - &
        - &
        - &
        89.11 &
        \textbf{89.67} &
        89.59 &
        88.90 &
        \cl{58.47}\\
        \hline
 \end{tabular}}

\end{table}

\begin{table}
\caption{Tag-wise Macro F1 score for the XLM-R model}
\label{tab:tagwise}
\centering
\begin{tabular}{|c| c | c |c|c|c|} 
 \hline
 {} & PERSON & LOC & ORG & MISC & OUTSIDE \\ [0.5ex] 
 \hline
 Sinhala & 97.45  & 82.23 & 82.65 & 83.23 &96.14 \\ 
 \hline
 Tamil & 79.35 & 81.43 & 69.49 & 76.88 &94.02 \\ 
 \hline
 English & 96.6 & 86.18 & 82.92 & 85.57 &96.71 \\ 
 \hline
\end{tabular}
\end{table}

Table \ref{tab:tb5} contains the final results. Table~\ref{tab:tagwise} shows the tag-wise distribution of results for the XLM-R model.

Recall that mBERT and IndicBERT have not been pre-trained for Sinhala, and SinBERT has not been pre-trained for Tamil or English. Nevertheless, we tested all these models for all the languages. 
 When a language is not included in the considered pLM, the corresponding result is grayed out.

First and foremost, we note that Bi-LSTM CRF results with LSTM-generated input embeddings and prefix features performed the best among the Bi-LSTM models. However, these results are far below that of pLMs, when the considered language is included in the pLM. Note that we did not generate Bi-LSTM CRF results for English, because the pLMs reported the highest results for English. 

However, compared to the Bi-LSTM CRF results, pLMs that have not been pre-trained for a particular language show inferior results for that considered language (e.g.~mBERT results for Sinhala). On the positive side, SinBERT, which was trained with just 15.7 million Sinhala sentences, significantly outperforms the corresponding Bi-LSTM CRF results. This is a promising sign for other low-resource languages as well. Given that many languages have monolingual data, a language-specific pLM can be easily trained and fine-tuned for downstream tasks such as NER.

Similar to Conneau's~\cite{bib17} observations, mBERT performs  slightly worse than XLM-R for both English and Tamil. On the other hand, IndicBERT falls below the two mLMs. This drop is significant for Tamil. This is very surprising given that IndicBERT was specifically trained on Indic languages, including Tamil and some other languages from the Dravidian family. However, we also note that in Kakwani et al.'s~\cite{kakwani2020indicnlpsuite} results, IndicBERT falls behind XLM-R and/or mBERT in several NLP tasks. Another interesting fact is SinBERT result for Sinhala, which falls behind XLM-R. A similar observation was made by Dhananjaya et al.~\cite{dhananjaya2022bertifying} for sentiment analysis. We believe the superior performance of XLM-R is due to its cross-lingual transfer abilities~\citep{asai2023buffet}. This confirms that mLMs can be a better alternative to language-specific pLMs, specifically when the amount of language-specific data is limited to train pLMs.

There is a noticeable variance across the XLM-R results for the three languages. Given that this is a multi-way parallel dataset (i.e. the same dataset got annotated across the three languages), this variance can be explained with two factors: representation of individual languages in the mLM, and complexity of individual languages. For example, English, which has the highest representation in XLM-R (i.e. English language data has been used much more than other languages during XLM-R model pre-training) has shown the highest performance. However, despite Tamil having more representation in XLM-R than Sinhala, XLM-R result for Tamil is much lower than Sinhala. This could be due to the language complexity (Tamil is  more agglutinative than Sinhala). Another reason could be Dravidian languages (where Tamil belongs to) being under-represented in XLM-R compared to the Indo-Aryan languages (where Sinhala belongs to)~\citep{ranathunga2022some}.

Finally, we note that multilingually fine-tuned XLM-R (mXLM-R) performs the best for both Sinhala and Tamil. However, there is a slight drop in the result for English. We believe this is due to the phenomenon known as negative interference, which states that the model performance for high-resource languages drops in multilingual settings~\citep{wang2020negative}. Nevertheless, since this single model can handle all three languages, this multilingual model is the best pick for our task.


\section{Case Study - Improving Neural Machine Translation with NER}
Named Entity translation is challenging even to the modern day NMT systems~\citep{laubli2020set}. Early solutions to this problem involve the use of external transliteration systems~\citep{grundkiewicz2018neural,ameur2017arabic,zhang2020translation}. However, such NE transliteration systems focus on translating individual NEs with no regard to the context of the NEs, which makes it difficult to resolve ambiguities. This is particularly problematic in highly inflected languages such as Sinhala and Tamil. Hu et al.~\cite{hu2021deep} recently proposed an alternative solution that involves pre-training a language model prior to fine-tuning it for NMT. Since their results seem to far exceed the results of NMT models that do not have explicit NE translation capabilities, we selected this model as the case study to demonstrate the usability of our NER system.  



\begin{figure}[!h]
\begin{center}
\includegraphics[scale=0.5]{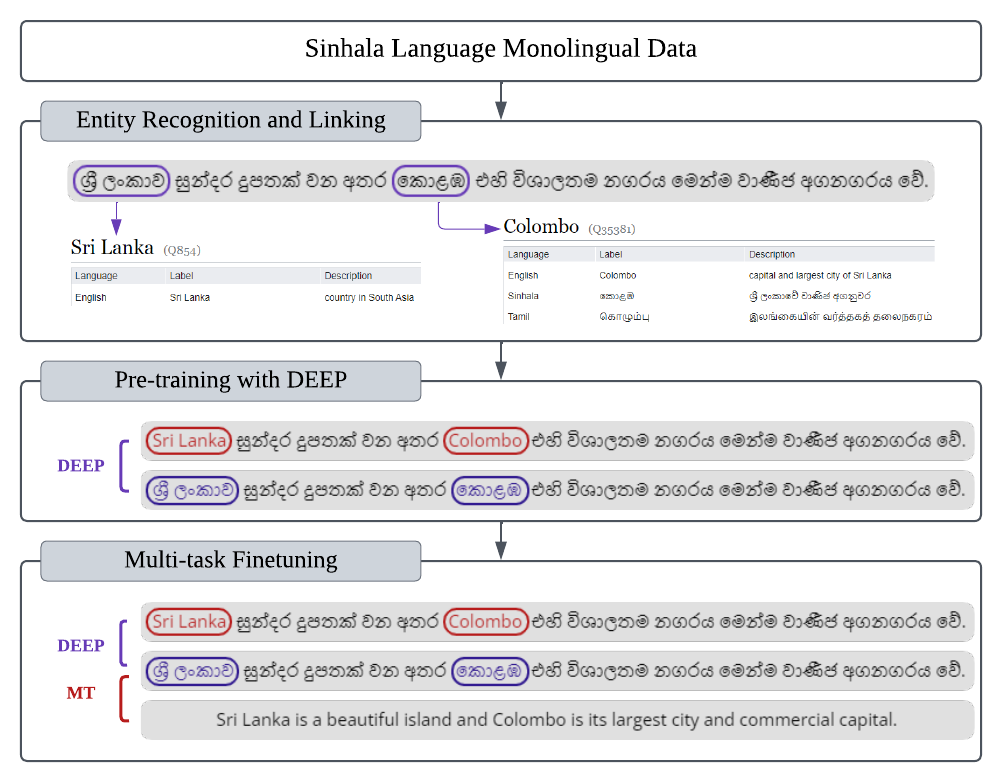} 
\caption{DEEP\citep{hu2021deep} Architecture}
\label{fig.4}
\end{center}
\end{figure}

Hu et al.~\cite{hu2021deep} adopted a pre-training and fine-tuning process to implement an NMT system that has better capabilities for NE translation. They first identified NEs in a monolingual corpus and linked them to a knowledge base (KB) that contains entity translations (WikiData~\citep{vrandevcic2014wikidata} in their case). For NE linking, they used the SLING \citep{ringgaard2017sling} entity linker. Then the entity translations in the KB were used to generate noisy code-switched data. After that, a Transformer model~\citep{vaswani2017attention} was pre-trained with this noisy code-switched data using the de-noising pre-training objective. Note that this is the same objective that was used to train the encoder-decoder pLM mBART~\citep{tang2021multilingual}. In de-noising pre-training, the Transformer model is taught to reconstruct an original sentence from its noised version. Finally to further improve translation of low-frequency NEs, they used a multi-task learning strategy that fine-tunes the
model using both the denoising task on the monolingual data and the translation task on the parallel
data. This model, known as DEEP (DEnoising Entity Pre-training) is illustrated in Fig.~\ref{fig.4} using English and Sinhala as the example (note the code-switched data used for pre-training). 

We first implemented Hu et al.'s~\cite{hu2021deep} DEEP NMT system using SLING, as well as a baseline NMT system without DEEP, for English-Sinhala (En-Si) translation. We used WikiData~\citep{vrandevcic2014wikidata} as the monolingual dataset, and Fernando et al.'s~\cite{fernando2020data} parallel data to train the NMT model. To elaborate further, the baseline NMT system was built by simply fine-tuning a Transformer model from scratch, similar to Hu et al.~\cite{hu2021deep}. To build the DEEP NMT system, first the NEs in the Sinhala WikiData were identified and linked to their corresponding English NEs using SLING. Then the NEs in the Sinhala WikiData were replaced with the identified English NEs. These code-switched noisy sentences were used to pre-train a Transformer model using the de-noising objective. Finally, this pre-trained Transformer model was further fine-tuned with the training set of Fernando et al.'s~\cite{fernando2020data} parallel data using the NMT objective.

However, as shown in Table~\ref{tab:tb6}, the DEEP system with SLING lags behind the baseline system with respect to the BLEU score. This is not surprising since the SLING entity linker used in DEEP does not support Sinhala, and the WikiData corpus that was used as the KB is rather small for Sinhala. 

Next, we re-implemented the DEEP system by replacing SLING with our best NER system. {In other words, we used our NER system to identify NEs in Sinhala WikiData. Entity linking across the languages was done using the Pywikibot library~\citep{pfundner2015utilizing}.} As shown in Table~\ref{tab:tb6}, the resulting NMT system significantly outperforms both baseline and DEEP with SLING (by about 9 BLEU points). 

Similar to Hu et al.~\cite{hu2021deep}, we also calculated entity translation accuracy. Here, we first count the number of NEs in the target side (i.e.~Sinhala) of the test set used to test the NMT system. Out of these NEs, the number of NEs that got correctly translated by the NMT system is taken as the entity translation accuracy.  As shown in Table~\ref{tab:tb6}, the NMT model that incorporates our NER output shows the highest NE translation accuracy. 

\begin{table}[t]
    \caption{Named entity translation results on En-Si.}
    \label{tab:tb6}
    \centering
    \begin{tabular}{|c| c | c |} 
  \hline
        \bf{Technique} &
        \bf{BLEU} &
        \bf{Entity Translation Acc.} \\
        \hline
        NMT model without DEEP &
        11.9 & 
        49 \\
        \hline
        DEEP+SLING &
        11.59 & 
        47.94 \\
        \hline
        DEEP+NER+Wiki data linking &
        \textbf{21.12} & 
        \textbf{62.75} \\
        \hline
\end{tabular}

\end{table}

\section{Conclusion}
This paper presented a multi-way parallel Named Entity annotated dataset for Sinhala-English-Tamil. According to the best of our knowledge, this is the first multi-way parallel NE annotated corpus for any language pair. We carried out a comprehensive evaluation of different types of pre-trained pLM for the task of Named Entity Recognition. We present our multilingual NER model, which was trained with the multi-way parallel corpus as our best pick for the task. We also demonstrated the utility of our NER system, by plugging it into an English-Sinhala NMT model. In future, we plan to increase the size of the dataset further and extend the annotation to cover more NEs. 

\section*{Acknowledgements}
We thank Rameela Azeez for her initial involvement in the project.

\section*{Declaration}
This research did not receive any specific grant from funding agencies in the public, commercial, or not-for-profit sectors. Generative AI tools were not used to prepare the manuscript. The authors do not have anything else to declare.
  \bibliographystyle{elsarticle-num} 
  \bibliography{bibliography}



\end{document}